\documentclass{article}

\usepackage{fullpage}
\usepackage{setspace}
\usepackage{color}
\usepackage{graphicx}
\usepackage{float}
\usepackage{array}
\usepackage[countmax]{subfloat}

\usepackage{amsmath,amsfonts,amssymb}

\newcommand{\etal}{{et~al. }}
\newcommand{\pauli}[1]{{\sigma_{{#1}}}}
\newcommand{\jacobi}{{\mathcal{J}}}

\newcommand{\Real}{{\mathbb R}}
\newcommand{\Complex}{{\mathbb C}}
\newcommand{\CPerp}{{\mathbb C}^{\perp}}

\newtheorem{thrm}{Theorem}
\newtheorem{conj}[thrm]{Conjecture}

\begin{document}


\title{A Ternary Non-Commutative Latent Factor
    Model for Scalable Three-Way Real Tensor Completion}
\date{First online version: 26 October 2014}

\makeatletter

\author{G. Baruch, \texttt{Yahoo Labs, guy.baruch@icloud.com}}

\makeatother

\maketitle

\begin{abstract}
Motivated by large-scale Collaborative-Filtering applications, we present a
Non-Commuting Latent Factor (NCLF) tensor-completion approach for modeling
three-way arrays, which is diagonal like the standard PARAFAC, but wherein
different terms distinguish different kinds of three-way relations of
co-clusters, as determined by permutations of latent factors.

The first key component of the algebraic representation is the usage of two
non-commutative real trilinear operations as the building blocks of the
approximation. These operations are the standard three dimensional
triple-product and a trilinear product on a two-dimensional real vector
space~$\CPerp\subset \Real^{2\times2}$, which is a representation of the real
Clifford Algebra~$Cl(1,1)$ (a certain Majorana spinor).  %
Both operations are purely ternary in that they cannot be decomposed
into two group-operations on the relevant spaces.
The second key component of the method is combining these operations using
permutation-symmetry preserving linear combinations.

We apply the model to the MovieLens and Fannie Mae datasets, and find that it
outperforms the PARAFAC model.
We propose some future directions, such as unsupervised-learning.

\end{abstract}

\section{\label{sec:intro}Introduction}

Tensor completion of three-way arrays%
\footnote{
        Semantics of the term ``tensor'' differs between research
        communities, as elucidated in Section 2 of~\cite{desilva-2008}.
        We will take ``tensor'' to be equivalent of ``n-way array''. %
}
had been used to model three-way interactions in many experimental fields,
starting in the 1920s with the chemometrics and psychometrics communities.
Kolda and Bader provide an extensive review of tensor factorization
literature up to 2009~\cite{kolda-review-2009}.
A shorter but more current review is given by Graesdyck \etal 
in~\cite{grasedyck-survey-2013}. 

\begin{figure}
    \begin{center}
        \includegraphics[clip,width=0.8\textwidth]{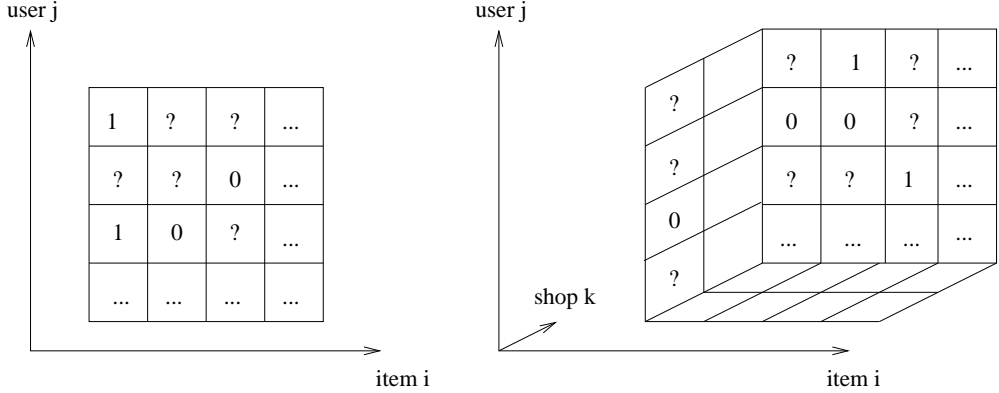} 
    \caption{\label{fig:mat_tensor}
    Left: An example of the classical Boolean Collaborative Filtering (CF)
    problem, wherein a binary response variable (indicating for example a
    purchase event) is given for each user-item pair, represented as a sparse
    matrix. Questions marks denote unknown values.
    The problem is estimating the probability of a purchase events for an unseen
    pair.
    Right: The corresponding three-way CF problem we consider, where the
    response variable depends on a triplet, in this case of user, item and 
    shop, represented as a sparse cuboid.
    }
    \end{center}
\end{figure}
This work considers three-way interactions in a ``Collaborative Filtering''
(CF) context. 
In the classical CF problem, some quantity of interest~$M$ (deterministic or
stochastic) depends on two variables of large cardinality~$M=M(i,j)=M_{ij}$ where $i=1,\dots,I$
and~$j=1,\dots,J$, which is naturally represented as a matrix.
The matrix of known values is typically sparse, and the problem is to estimate
the missing values, seeking the best approximation in the~$L^2$ (Froebenius)
norm.
In the three-way case the quantity of interest depends on three variables and is
represented as a cuboid tensor~$T=T(i,j,k)=T_{ijk}$.
See Figure~\ref{fig:mat_tensor} for an illustration and Section~\ref{sec:Formulation_3WCF} 
for a more concrete example.

The two main tensor decompositions used are the CANDECOMP/PARAFAC (CP) model 
proposed by Hitchcock in
1927~\cite{hitchcock-rank-1927,hitchcock-sum-1927},
and the Tucker decomposition proposed by Tucker in
1963~\cite{Tucker,tucker-factor-1963,tucker-factor-1964}.
In the CP model, a three-way array~$T\in\mathbb{R}^{I\times J\times K}$
is approximated by a finite sum of rank-1 tensors
\begin{equation}
    T_{ijk}^{\text{CP}} = \sum_{r=1}^{R}U_{ir}V_{jr}W_{kr}+\text{bias terms},\label{eq:CP}
\end{equation}
where $U\in\mathbb{R}^{I\times R}$, $V\in\mathbb{R}^{J\times R}$,
$W\in\mathbb{R}^{K\times R}$ are called ``latent factor matrices'', and
\[
    \text{bias terms}=b_{0}+b_{1i}+b_{2j}+b_{3k}.
\]
For readers unfamiliar with machine learning terminology, we note that the
name ``latent factor'' stems from an assumption that the data is generated
from a fixed distribution governed by variables which are hidden (latent).

In the more general Tucker model the latent factor rows are multiplied
by a ``core tensor''~$C_{R_1\times R_2\times R_3}$ of dimensions~$R_1,R_2,R_3$ as 
\begin{equation} 
    T_{ijk}^{\text{TUCKER}} = 
    \sum_{r_1=1}^{R_1}\sum_{r_2=1}^{R_2}\sum_{r_3=1}^{R_3}
        C_{r_1r_2r_3}U_{ir_1}V_{jr_2}W_{kr_3}
        +\text{bias terms}.\label{eq:tucker}
\end{equation}
The Tucker model is more expressive than the CP model, but its core tensor is
typically dense, requiring~${\mathcal{O}}\left( R^3 \right)$ parameters.
It is also harder to interpret.

We note that the CP model has the property that its basic building block - the
real triple product~$u_iv_jw_k$  - does not distinguish between cases wherein the
numerical values of the latent factors are permuted, for example between~$
(u_i,v_j,w_k)=(1,2,3) $ and~$ (u_i,v_j,w_k)=(2,3,1) $
(and similarly for other permutations).
In other words, for the three-way interactions modeled by CF, a commutative
building block is inherently less expressive than a non-commutative one.
Thus, we speculate that three-way relations are better distinguished by a
product of non-commuting latent factors than by the (commutative) real
multiplication of the CP model.  This intuition is expanded in
Section~\ref{sec:intuitive}.

Following this speculation, we propose a hybrid of the CP and
the Tucker3 models which is pseudo-diagonal (like the CP), but is built 
ground-up from trilinear operations of Non-Commuting Latent Factors (NCLF).
The general form of the NCLF model is 
\begin{equation}
    \begin{gathered}
    T_{ijk}^{\text{NCLF}} = 
    \sum_{\text{sym}=1}^{6}
    \sum_{r=1}^{R_{\text{sym}}}
    L_{\text{sym}}\left(
        U^{\text{sym}}_{ir}, 
        V^{\text{sym}}_{jr}, 
        W^{\text{sym}}_{kr}
    \right)
        +\text{bias terms}, \\
        U^{\text{sym}} \in \mathbb{V}^{I\times R_{\text{sym}}},\quad
        V^{\text{sym}} \in \mathbb{V}^{J\times R_{\text{sym}}},\quad
        W^{\text{sym}} \in \mathbb{V}^{K\times R_{\text{sym}}}, 
    \end{gathered}
        \label{eq:NCLF1}
\end{equation}
where the subscript ``sym'' denotes different permutation symmetries of latent factors,
$L_{\text{sym}}(\cdot)$ is a real trilinear mapping satisfying this symmetry mode,
and~$\mathbb{V}$ is a real linear space to be determined.

A well-known problem of unregularized CP models is that approximations of a certain
rank may not exist, a situation commonly called ``degeneracy'',
see Section~3.3 of~\cite{kolda-review-2009} and 
also~\cite{ComonLuciani-TdAlsEtc-2009}.
De Silva and Lek Heng Lim show that such degeneracy can be generic, i.e.,
occurring at a non zero-measure set of inputs~\cite{desilva-2008}.
They also prove that degeneracy always co-occurs with the formation of 
collinear columns of the latent factor matrices, meaning that the set of vectors~$
    U_{:1}, U_{:2},\dots, U_{:r}
$, where the colon sign denotes a running index, becomes linearly dependent,
or almost so. This dependency manifests in very large columns which almost
cancel each other.
They also note that, while regularization removes non-existence, proximity of
the well-posed regularized problem to the ill-posed unregularized problem may
still result in catastrophic ill-conditioning.

Much of the effort in lower-dimension tensor factorization have been
directed into extending the Singular Value Decomposition (SVD), for example
by applying orthogonality constraints on the columns of the latent factor
matrices or of the core matrix of the Tucker decomposition - see a review
in~\cite{kolda-ortho-2001}.
Orthogonality of matrix-slices of the Tucker core tensor has been considered by
L. de Lathauwer et al., who show that this model retains many properties of the
original matrix SVD, therefore naming it the High Order SVD
(HOSVD)~\cite{lathauwer-svd-2000}. 
The core tensor, however, is still dense
requiring~${\mathcal{O}}\left( R^3 \right)$ parameters.

When the dimension of the factors is small, orthogonality and collinearity of
the latent matrix columns are mutually exclusive, and orthogonality removes
degeneracy even for the CP model.
For typical ``big data'' CF problems, however, dimensionality of each factor
may be extremely large%
\footnote{
    For example, each Yahoo user may receive her own latent row vector,
    and the number of such users is in the hundreds of millions.%
} and so virtually all vector pairs are near-orthogonal.
Near-orthogonality is therefore not useful in avoiding collinearity.
We note that a standard CP expansion of a finite-rank NCLF model will always
have collinear parallel factors.
Hence, some degenerate modes may be alleviated by the NCLF model.
We leave the question of how much degeneracy is alleviated open%
\footnote{
    Some examples wherein the CP model becomes degenerate are associated
    with differential operators, see~\cite{desilva-2008}.
    The NCLF model directly models CP-degenerate modes associated
    with first-order finite-difference operators. 
    Therefore, we speculate it removes degeneracy associated with first-order
    differential operators, but not all the higher-order ones.
}.

In the completely different setting of particle physics, modeling three-way
interactions (in three-quark models) have been shown to be intrinsically related
to non-commutativity of the underlying algebras.
Kerner proposed using one such algebra in three-color quark
models~\cite{kerner-cubic-lorentz-2010}, and we shall use such ideas for
the algebraic representation used by our model%
\footnote{For the reader unfamiliar with physics we note that the CF problems we
    consider are entirely different from quantum chromodynamics, so that we can
    propose much simpler models.%
}.

For the reader familiar with Geometric Algebra we add two notes, which other
readers may safely ignore.
First, we will use the two dimensional real representation of the Clifford Algebra~$Cl(1,1)$,
which in Physics is known as one of the flavors of a Majorana spinor.
Second, some recent tensor factorization works use Grassman algebras to represent
the completely antisymmetric components of the
input~\cite{kolda-review-2009,kressner-riemanian-opt-2013}.
In the third order case the standard triple product
in~$\mathbb{R}^{3}$, which is the approach we use for this component,
is a Grassman Algebra.

The remainder of the paper is as follows.
In Section~\ref{sec:Formulation_3WCF} we formulate the specific CF problem we
are interested in.
In Section~\ref{sec:intuitive} we give the motivating intuitions of this work.
Specifically, we conjecture that in order to distinguish between three-way
relations by a single term, an algebraic representation must be
non-commutative.
Moreover, it must model, either implicitly or explicitly, different permutation
symmetries of the latent factors.
Following these intuitions, in Section~\ref{sec:tensor_approximation} we
construct the NCLF model, which we construct in several steps:
\begin{enumerate}
    \item In Section~\ref{ssec:NxNxN} we recall the decomposition of a
        generic cubical tensor into its symmetry-preserving components.
        This decomposition is done via six linear operators.
    \item In Section~\ref{ssec:space_CPerp} we look for and find a
        non-commutative trilinear mapping~$\mu$ on a two-dimensional linear
        subspace~$\CPerp$ of~$\Real^{2\times2}$, which is the simplest such
        mapping we could devise. 
        This mapping is the key component of our method, and will be used to
        construct five of the six symmetry-preserving components of the NCLF
        model.
        We denote this space by~$\CPerp$ because it is the orthogonal complement
        of the representation of the Complex field in~$\Real^{2\times2}$.
        The mapping~$\mu$ is purely ternary, meaning that the
        space~$\mathbb{C}^{\perp}$ is closed under the trilinear operation, but
        not under the corresponding bilinear one.
        In other words,~$\mathbb{C}^{\perp}$ is a ternary algebra, not a
        standard (binary) algebra.
    \item In Section~\ref{ssec:approximating_S_and_J} we approximate each of
        these components by its own trilinear mapping: the completely antisymmetric
        component is modeled by the standard triple-product in~$\mathbb{R}^{3}$,
        and approximation of the other components are constructed by applying
        the symmetrizing operations on the mapping~$\mu$.
        We provide explicit expressions for each of the components.
    \item Finally, in Section~\ref{ssec:general_cuboid} we assemble the full
        approximation, and apply it to the general cuboid case.
\end{enumerate}
In Section~\ref{sec:numerical}, we provide the results of numerical experiments
on two publicly available datasets, the MovieLens movie rating dataset and the
Fannie Mae Single Family Home Performance dataset.
In both cases, the non commutative models outperform the standard CP model.
We conclude and discuss future directions in Section~\ref{sec:discussion}.

\section{A specific Three-way CF problem }
\label{sec:Formulation_3WCF}

The specific problem motivating this paper is that of predicting binary response
via three-way CF in supervised learning.
In this learning problem, the dependent variable is a Boolean event - like a
purchase event, which we denote by~$Y\in\left\{ 0,1 \right\}$,
and the independent variables belong to three classes of large cardinality,
for example users, purchasable items and shopping venues, see
Figure~\ref{fig:mat_tensor} on the right.

The learning problem is therefore to estimate the 
probability of a purchase event~$P(Y=1|i,j,k)$ for an (unseen) triplet~$i\in1,\dots,I$,
$j\in1,\dots,J$ and $k\in1,\dots,K$.
The value of~$Y_{ijk}$ for most of the triplets is unknown,
making this a tensor completion problem.

We will use a Logistic Regression model, thereby estimating the
log-odds of this probability
\begin{subequations}
    \label{eqs:multilinear_log_reg}
    \begin{equation}
        \log\left( 
        \frac{P(Y=1|i,j,k)}{1-P(\cdot)}
        \right)
        \approx T_{ijk} \in \Real^{I\times J\times K},
    \end{equation}
    or, equivalently,
    \begin{equation}
        P(Y=1|i,j,k) \approx
        \text{Logit}\left( T_{ijk} \right),\qquad 
        \text{Logit}(T_{ijk}) = \frac1{1+\exp\left( -T_{ijk} \right)}.
    \end{equation}
\end{subequations}

We will be using~$L^2$ (Tikhonov) regularized models and the logistic loss
function, so that given a functional form~$T(U,V,W,b)$ (like CP, or Tucker3)
and data~$Y_{ijk}$ (known over a subset of the triplets~$(i,j,k)$), training
will consist of the solution of the minimization problem
\begin{gather}
    U,V,W,b = \arg\min
    \sum_{
        \left\{ 
            (i,j,k): Y_{ijk} \text{ known}
        \right\}
    }
    - Y_{ijk}\log\left( \text{Logit}(T_{ijk}) \right)
    - (1-Y_{ijk})\log\left( 1-\text{Logit}( T_{ijk}) \right)
    \nonumber \\
    + \lambda \left\Vert U \right\Vert^2
    + \lambda \left\Vert V \right\Vert^2
    + \lambda \left\Vert W \right\Vert^2,
    \label{eq:loss}
\end{gather}
where the last three terms are the regularization terms, and the
parameter~$\lambda$ is the regularization parameter, to be chosen empirically
via cross validation.

These four simplifying assumptions - of a supervised learning, binary response
problem modeled by logistic regression with~$L^2$ regularization - are applied in 
order to demonstrate the NCLF model on a concrete problem.
Apriori, they only affect the numerical experiments in
Section~\ref{sec:numerical}.
We see no reason why the NCLF model should not apply to other three-way
multilinear subspace learning problems.

\section{\label{sec:intuitive}The intuitive motivation}

Let us look for the simplest extension to the trilinear CP model,
which would still be be diagonal, but would provide a more expressive
algebraic representation of a three-way relation between entities,
for example between users, purchasable items and venues. 
Such a representation approximates how a three-way relation affects some
measured quantity - for example the odds~$P/(1-P)$ of a purchase event - which
we take for simplicity to be real.
Since we are estimating a real quantity, we consider real trilinear mappings.

Following intuitions from Physics~\cite{kerner-cubic-lorentz-2010}, we speculate
that non-commutative parallel factors might be more expressive than commutative
ones, i.e., that in reality a
``green user, blue item red shop'' combination is different than a
``blue user, green item, red shop'' combination, and will lead to a different
propensity to purchase.
Since the ``colors'' are arbitrary regions of the latent factor
space corresponding to different co-clusters, there is no reason, priory, to
assume that a function representing the relation between parameter regions
for shops, items and venues be commutative in the latent factors.

Hence, this article raises the following conjecture:
\begin{conj}\label{conj:noncom}
    A trilinear tensor completion model which is built upon non-commutative
    parallel factors, i.e., that differentiates between different
    permutations of the same numerical values of its arguments, would in some
    way be ``more realistic'' - hence perform better than the standard CP
    model.
\end{conj}

Conjecture~\ref{conj:noncom} leads to two immediate outcomes.
Firstly, the standard CP model is suboptimal - since its building block is the
multiplication of real arguments and is inherently commutative.
If a trilinear building block is to be used, the arguments must be of dimension
two at least.
Likewise, the next simplest extension which is the multiplications of complex
arguments, cannot be used (at least naively), as it is commutative.
Secondly, in order to differentiate between all different ``color''
permutations of three objects, there must be at least three ``colors''.
In other words, a single parallel factor must differentiate at least three
co-clusters of each class.
Non-commutative three-way relations between co-clusters must therefore
involve, at the very least, a~$3\times3\times3$
assignment - a mapping $
    \left\{ \text{red}, \text{green}, \text{blue}\right\}^3
    \mapsto \Real $.

In the next Section we construct such a real trilinear approximation of
three-way arrays in~$\Real^{N\times N\times N}$ for~$N\geq3$.
We shall later use this construction for a general tensor completion problem.

\section{The Non Commutative Latent Factors (NCLF) method}
\label{sec:tensor_approximation}

\subsection{\label{ssec:NxNxN}Approximating a real~$N\times N\times N$ array}

\begin{table}
\begin{centering}
\begin{tabular}{|c||c|c|c|c|c|c|}
\hline 
component & $P_{{\rm cyc}}$ & $P_{{\rm acyc}}$ & $P_{J}$ & $P_{12}$ & $P_{23}$ & $P_{31}$\tabularnewline
\hline 
\hline 
$S$ & $1$ & 1 & NA & $1$ & $1$ & $1$\tabularnewline
\hline 
$A$ & $1$ & $-1$ & NA & $-1$ & $-1$ & $-1$\tabularnewline
\hline 
$\jacobi_{31-}$ & NA & NA & $0$ & NA & NA & $-1$\tabularnewline
\hline 
$\jacobi_{31+}$ & NA & NA & $0$ & NA & NA & $1$\tabularnewline
\hline 
$\jacobi_{23-}$ & NA & NA & $0$ & NA & $-1$ & NA\tabularnewline
\hline 
$\jacobi_{23+}$ & NA & NA & $0$ & NA & $1$ & NA\tabularnewline
\hline 
\end{tabular}
\par\end{centering}

\caption{\label{tab:cuboid_symmetrizers}Eigenvalues of the components of a
cubical three-way array $T_{ijk}$ given in
eq.~\eqref{eq:cuboid_symmetrizers} under the generic cyclic and acyclic
permutation operators $P_{{\rm cyc}},\, P_{{\rm acyc}}$,
the Jacobi-like operator $J=T_{ijk}+T_{jki}+T_{kij}$, and the index-pair
exchange of~$i,j$ denoted by $P_{ij}$. }
\end{table}

We recall that, given a three-dimensional cuboid array of real numbers
$T\in {\mathbb R}^{N \times N \times N}$,
it may be decomposed to six components according to their permutation symmetry 
properties.
There are several options for doing this, and the decomposition we choose is
\begin{equation} \label{eq:cuboid_symmetrizers}
    \begin{bmatrix}
        S[T] \\
        A[T] \\
        \jacobi_{31-}[T] \\
        \jacobi_{31+}[T] \\
        \jacobi_{23-}[T] \\
        \jacobi_{23+}[T] \\
    \end{bmatrix}_{ijk}
    =
    \begin{bmatrix}
        1  &  1 &  1 &  1 &  1 &  1 \\
        1  &  1 &  1 & -1 & -1 & -1 \\
        1  &  0 & -1 &  1 &  0 & -1 \\
        1  &  0 & -1 & -1 &  0 & 1 \\
        0  &  1 & -1 &  0 & -1 & 1 \\
        0  &  1 & -1 &  0 &  1 & -1 
    \end{bmatrix}
    \begin{bmatrix}
        T_{ijk} \\
        T_{jki} \\
        T_{kij} \\
        T_{ikj} \\
        T_{jik} \\
        T_{kji}
    \end{bmatrix}.
\end{equation}
Eq~\eqref{eq:cuboid_symmetrizers} is a list of linear combinations of~$T_{ijk}$
and its index permutations.
We note that the linear mapping~\eqref{eq:cuboid_symmetrizers} is invertible and
well-conditioned.

The symmetry properties of the six components are given in 
Table~\ref{tab:cuboid_symmetrizers}.
The first two components~$S$ and~$A$ are eigenvectors of all the permutation
symmetries - the first being symmetric under all permutations while the
second being symmetric under cyclic (even) permutations and anti-symmetric
under acyclic (odd) ones.
The next four components are eigenvectors of only a single permutation symmetry
each, but all satisfy a Jacobi-like identity:
\begin{equation}
    \jacobi[T] := T_{ijk} + T_{jki} + T_{kij} \equiv 0.
    \label{eq:Bianchi}
\end{equation}

We use the images of these operators to define three
linear subspaces of ${\mathbb R}^{N \times N \times N}$.
The first two are the images of the totally symmetric and totally antisymmetric
operators~$Im(S)$ and~$Im(A)$.
The third subspace is the sum of the images of the last four operators, which
is also equal to the kernel of the Jacobi identity~$
Im\jacobi_{23+} +
Im\jacobi_{23-} +
Im\jacobi_{31+} +
Im\jacobi_{31-}
= Ker(\jacobi)
$.
Direct calculation gives that, taken as subspaces 
of~$\Real^{N \times N \times N}$ with the Euclidean inner product
associated with the Froebenius norm, the three spaces are pairwise orthogonal
and span the full space, hence
$ \Real^{N \times N \times N} = Im(A)\oplus Im(S) \oplus Ker(J) $.

Next, we construct diagonal trilinear approximations of for each of these
six components, which satisfy the relevant symmetries.
The second component~$A[T]$ is approximated using the standard totally
antisymmetric form, or standard triple product in $\Real^3$, which is equal
to~$\text{det}\left[ u~v~w \right]=u(v\times w)$, with three-dimensional latent
factors~$u,v,w\in\Real^3$.
In the next two Sections, we approximate the other five components using a
two-step process:
\begin{enumerate}
    \item In Section~\ref{ssec:space_CPerp} we define a trilinear
        non-commutative mapping, which we shall denote by~$\mu$,
        over a two-dimensional subspace of~$\Real^{2\times2}$.
        As it is two dimensional, it is hard to think of a simpler such mapping.
    \item Next, in Section~\ref{ssec:approximating_S_and_J} we apply the
        symmetrizing operators of~\eqref{eq:cuboid_symmetrizers} on this
        trilinear form~$\mu$, to obtain the approximations for the five
        components.
\end{enumerate}
In Section~\ref{sec:numerical} we provide numerical indications that each of 
these two steps improves the overall approximation of the chosen
datasets.

\subsection{The space~$\CPerp$ and operation~$\mu$ }
\label{ssec:space_CPerp}

Let us look for the simplest ``atom'' for the Jacobi components - that is the
simplest possible space supporting a noncommutative trilinear product.
This space is the key component of our mathematical model.
We note that the complex version of this space has been used in computational
Physics of three-color quantum models~\cite{kerner-cubic-lorentz-2010}.

A trilinear operation with one dimensional real arguments must be commutative, and
so such a space must have at least two dimensional arguments.
Non-commutativity and trilinearity leads us towards~$2\times2$ matrix multiplication
as a representation.

Before we continue, let us recall two basic facts on the space of~$2\times2$
real matrices ~$\Real^{2\times2}$.
First, it is spanned by the identity matrix and the three Pauli spin matrices:
\[
    \mathbf{I}=\left[\begin{array}{cc}
            1 & 0\\
            0 & 1
    \end{array}\right],
    \qquad
    \pauli{1}=\left[\begin{array}{cc}
            0 & 1\\
            1 & 0
    \end{array}\right],
    \qquad
    i\pauli{2}=\left[\begin{array}{cc}
            0 & -1\\
            1 & 0
    \end{array}\right],
    \qquad
    \pauli{3}=\left[\begin{array}{cc}
            1 & 0\\
            0 & -1
    \end{array}\right],
\]
which are mutually orthogonal in the inner product associated with the Froebenius
norm. In other words they are an orthogonal basis of~$\Real^{2\times2}$.
Second, the space of complex numbers~$\mathbb C$ is isomorphic, using the
Cayley-Dickson construction, to the space of antisymmetric~$2\times2$ real
matrices of the form
\[
    \text{CD}(\Complex)
    = \left\{ 
        \left.
        \left[\begin{array}{cc}
                c^0 & c^2 \\
                -c^2 & c^0
        \end{array}\right]
        \right|
        c^0, c^2 \in\mathbb{R}
    \right\}
    = \left\{
        c^0\mathbf{I} + c^2i\pauli{2} \Big| c^0, c^2 \in\mathbb{R}
    \right\}
\]
with matrix multiplication corresponding to the product of complex numbers.
In this subspace of~$\Real^{2\times2}$, matrix multiplication is commutative.

With these facts in mind, we therefore turn to the orthogonal
complement~$\CPerp$ of~$\text{CD}(\Complex)$ to look for
non-commutative trilinear operations.
From the fact that~$\left\{
    \mathbf{I}, \pauli{1}, \pauli{2}, \pauli{3}
\right\}$ is an orthogonal basis it immediately follows 
that~$\CPerp$ is the span of~$\left\{ \pauli{1},\pauli{3}\right\}$:
\begin{equation}    \label{eq:V_B}
    \CPerp := \left\{
        c^1\pauli{1} + c^3\pauli{3} \Big| c^1,c^3 \in\mathbb{R}
    \right\}
    = \left\{ 
        \left.
        \left[\begin{array}{cc}
                c^3 & c^1\\
                c^1 & -c^3
        \end{array}\right]
        \right|
        c_1,c_3 \in\mathbb{R}
    \right\}
    .
\end{equation}
It is also the space of traceless symmetric~$2\times2$ real matrices.

Additionally, for each ordered triplet~$u,v,w\in\CPerp$, setting
\begin{equation} \label{eq:v_i}
    u=u^{1}\pauli{1}+u^{3}\pauli{3},
\end{equation}
and similarly for~$v,w$, direct calculation shows that~$\CPerp$ is closed
under a triple matrix product:
\begin{eqnarray*}
    uvw
    & = & \left(
        u^1v^1w^1+u^3v^3w^1-u^3v^1w^3+u^1v^3w^3
    \right)\pauli{1}\\
    &  & +\left(
        u^3v^3w^3+u^1v^1w^3-u^1v^3w^1+u^3v^1w^1
        \right)\pauli{3}.
\end{eqnarray*}
Hence, the mapping
\begin{equation}
    \begin{array}{ccc}
        \mu:\CPerp\times \CPerp \times \CPerp &\to& \CPerp \\
        \mu(u,v,w) &\mapsto& uvw
    \end{array}
    \label{eq:bianch_operator}
\end{equation}
is a well defined real trilinear operation.
Considering commutativity, the product~$uvw$ is symmetric with
respect to exchange of the first and third parameters, but not to a permutation which
changes the second argument%
\footnote{
    Indeed, the algebra~$Cl(1,1)$ is defined as the two dimensional Clifford
    Algebra having one symmetric and one antisymmetric index.
}
\begin{equation}
    \label{eq:Cperp_symmetry}
    uvw = wvu, \qquad uvw\neq uwv,\qquad
    u,v,w\in\CPerp.
\end{equation}

We note that~$\CPerp$ is not closed under the standard (binary) matrix
multiplication - for~$u,v\in\CPerp$ we have~$uv\in CD(\Complex)$,
not~$\CPerp$.
Therefore,~$\CPerp$  is not a group under matrix multiplication, and is hence
not an algebra, but rather a ternary algebra.
Similarly to the standard triple product in $\Real^3$, the
pair~$(\CPerp, \mu)$ is a purely third-order construct.

\subsection{Approximating the five components}
\label{ssec:approximating_S_and_J}

Here, we approximate the symmetric and Jacobi components
of~$T_{ijk}$, which are~$S$ and~$\jacobi_{31\pm}, \jacobi_{23\pm}$,
using linear combinations of the form~$\mu$ on~$\CPerp$.
Specifically, if the latent factor corresponding to item~$i$ is \[
    u_i = u_i^1\pauli{1} + u_i^3\pauli{3} \in \CPerp, 
\]
and similarly for~$v_j$ and~$w_k$, we apply the symmetrizing operators
of~eq.~\eqref{eq:cuboid_symmetrizers} on~$\mu(u_i,v_j,w_k)$ to obtain
these operators as explicit cubic polynomials of the coefficients.
For example, the totally symmetric component is
\begin{subequations}
    \label{eqs:explicit_S_und_J}
    \begin{eqnarray}
        S\left(u_i,v_j,w_k\right) &:=& \nonumber
            u_iv_jw_k +v_jw_ku_i +w_ku_iv_j
            +u_iw_kv_j +w_kv_ju_i +v_ju_iv_j \\
        &=& 2\left(
            3u_i^1v_j^1w_k^1 +u_i^3v_j^1w_k^1 +u_i^1v_j^3w_k^1 +u_i^1v_j^1w_k^3
            \right) \pauli{1} \nonumber \\
        &&  + 2\left(
            3u_i^3v_j^3w_k^3 +u_i^1v_j^3w_k^3 +u_i^3v_j^1w_k^3 +u_i^3v_j^3w_k^1
            \right) \pauli{3},
    \end{eqnarray}
    and similarly 
    \begin{eqnarray}
        \jacobi_{31-}\left(u_i,v_j,w_k\right)
        &=& 2\left( u_i^1v_j^3w_k^3 - u_i^3v_j^3w_k^1 \right) \pauli{1}
        \nonumber \\
         && +  2\left( u_i^3v_j^1w_k^1 - u_i^1v_j^1w_k^3 \right) \pauli{3} 
         \\
        \jacobi_{31+}\left(u_i,v_j,w_k\right)
        &=& 2\left(
            u_i^1v_j^3w_k^3 -2u_i^3v_j^1w_k^3 +u_i^3v_j^3w_k^1
            \right) \pauli{1} \nonumber \\
          && + 2\left(
            u_i^3v_j^1w_k^1 -2u_i^1v_j^3w_k^1 +u_i^1v_j^1w_k^3
            \right)\pauli{3}
        \\
        \jacobi_{23-}\left(u_i,v_j,w_k\right)
        &=& 2\left( u_i^3v_j^3w_k^1 - u_i^3v_j^1w_k^3 \right) \pauli{1} 
        \nonumber \\
         && +  2\left( u_i^1v_j^1w_k^3 - u_i^1v_j^3w_k^1 \right) \pauli{3} 
         \\
        \jacobi_{23+}\left(u_i,v_j,w_k\right) 
        &=& 2\left(
            -2u_i^1v_j^3w_k^3 +u_i^3v_j^1w_k^3 +u_i^3v_j^3w_k^1
            \right) \pauli{1} \nonumber \\
         && + 2\left(
            -2u_i^3v_j^1w_k^1 +u_i^1v_j^3w_k^1 +u_i^1v_j^1w_k^3
            \right)\pauli{3}.
    \end{eqnarray}
\end{subequations}

Importantly, the symmetry~\eqref{eq:Cperp_symmetry} of~$\mu$ implies that the
completely anti-symmetric combination
vanishes \[
    A(u_i,v_j,w_k) := u_iv_jw_k + v_jw_ku_i + w_ku_iv_j - (w_kv_ju_i + u_iw_kv_j
    + v_ju_iw_k) \equiv 0.
\]
This is reassuring, as the Jacobi and symmetric components are orthogonal to the
anti-symmetric component.

\subsection{The general cuboid case}
\label{ssec:general_cuboid}
The previous subsections dealt with a cubical
array in~${\mathbb R}^{N\times N\times N}, N\geq3$.
We shall reuse the same model in the general cuboid case as is, without any
formal justification.
The intuition behind this is that the previous derivation applies to modeling
the relations of co-clusters (aka ``colors''), which can be cubical even if
the approximated tensor is a cuboid.
The ultimate judge is, of course, empirical evidence.

Therefore, combining the results of this Section, given a three-dimensional
(cuboid) array of real numbers~$T\in {\mathbb R}^{I \times J \times K}$,
we approximate it as 
\begin{eqnarray} \label{eq:overall_approx}
    T_{ijk}^{NCLF} &=& 
    b_0 + b_{1i} + b_{2j} + b_{3k} +
    \sum_{r=1}^{R_s} \zeta_r^S
    S\left( u_{ir}^{S}, v_{jr}^{S}, w_{kr}^{S} \right)
    + \sum_{r=1}^{R_A} \alpha_r
    \text{det}\left[ u^{A}_{ir}, v^{A}_{jr}, w^{A}_{kr} \right]
    \nonumber \\
    && +
    \sum_{p\in 23,31} \sum_{s=\pm}
    \sum_{r=1}^{R_{ps}}\zeta_r^{(ps)}
    \jacobi_{ps}\left(
    u_{ir}^{(ps)}, v_{jr}^{(ps)}, w_{kr}^{(ps)}
    \right),
    \nonumber \\
    && 
    (\cdot)^{S},~(\cdot)^{(ps)}\in\CPerp, 
    \quad (\cdot)^{A}\in\Real^3,
\end{eqnarray}
where~$b_{(\cdot)}$ are corresponding bias terms, the
operators~$S,\jacobi_{ps}$ are as defined in~\eqref{eqs:explicit_S_und_J},
$\det\left[ \cdot \right]$~is the standard triple product in~$\Real^3$
and the quantities~$\zeta_r^{(\cdot)}\in\Real^2$, which generalize singular
values, imply summation over the~$\pauli{1}$ and~$\pauli{3}$ components.

Equation~\eqref{eq:overall_approx} is the concrete, explicit model of the
general form~\eqref{eq:NCLF1}, and is the key result of this paper. 
Note that this approximation is as close to diagonal as possible, while still
being noncommutative, i.e., while differentiating between different
permutations of the latent factors, as required by
Conjecture~\ref{conj:noncom}.

\section{\label{sec:numerical}Numerical Experiments}

Here we present the results of numerical experiments for two public datasets -
the MovieLens Dataset~\cite{MovieLens-Dataset} and the Fannie Mae Single-Family
Loan Performance dataset~\cite{FannieMae-DataSet}.
The goal of experiments was a comparison of the 
expansion~\eqref{eq:overall_approx} with the standard CP model,
rather than obtaining the optimal model for each Dataset.
In both cases we used a binary response variable and a logistic-regression
model, so that the probability of a positive event is modeled
by~\eqref{eqs:multilinear_log_reg} and training consists of solving the
minimization problem~\eqref{eq:loss}, see Section~\ref{sec:Formulation_3WCF}.

\subsection{\label{ssec:approximations}Benchmark Approximations}
Five benchmark approximations of the logodds~$T_{ijk}$ were compared:
\begin{enumerate}
    \item A bias-only method, which is equivalent to a Naive Bayes
        approximation:
        \[
            T_{ijk}^{\text{Bias only}} = b_0 + b_{1i} + b_{2j} + b_{3k}.
        \]
        The total logodds bias~$b_0$ and the relative biases~$b_{fi}$ for each
        entity~$i$ of factor~$f=1,2,3$ were estimated as empirical logodds
        \begin{equation} \label{eq:bias_terms} 
            b_0=\log\frac{P+1}{N+1},
            \qquad b_{fi}=\log\frac{P_{fi}+1}{N_{fi}+1}-b_0.
        \end{equation}
        where~$P,N$ are the total counts of positive and negative events for the
        training set and~$P_{fi},N_{fi}$ are the same counts for each
        entity~$fi$.
    \item The standard CP approximation~\eqref{eq:CP} with a latent dimension
        equal to that of the NCLF method~$R=13$.
    \item The standard CP with the best latent dimensions~$R=5$ for both the
        MovieLens and Fannie Mae datasets.
        The best dimensions were chosen via nine-fold cross-validation.
    \item In order to test the utility of  the derivation of 
        Section~\ref{ssec:approximating_S_and_J}, i.e., of 
        using the separate approximations~\eqref{eqs:explicit_S_und_J}
        for each of the five components~$S$ and~$\jacobi_{(\cdot)}$,
        we also benchmark a ``primitive'' NCLF approximation given by
        \begin{equation} \label{eq:mu_only_parafac}
            \begin{gathered}
            T^{{\rm primitive~NCLF}}_{ijk} =
            b_0 + b_{1i} + b_{2j} + b_{3k}
            + \text{det}\left[ u^{A}_{ir}, v^{A}_{jr}, w^{A}_{kr} \right] \\
            + \sum_{r=1}^{5}\zeta_{r}\mu\left( 
                u_{ri}, v_{rj}, w_{rk}
            \right).
            \end{gathered}
        \end{equation}
        This approximation explicitly models only the totally-antisymmetric
        component~$A$, while using the primitive operation~$\mu$ instead of
        modeling each of the five components~$S$ and~$\jacobi_{(\cdot)}$.
        We recall that~$\mu$ has partial symmetry~\eqref{eq:Cperp_symmetry}.
        This implies that the partially-antisymmetric
        components~$\jacobi_{23-},\jacobi_{31-}$ are not approximated
        by~\eqref{eq:mu_only_parafac}, while the rest of the components are.
    \item The proposed NCLF method, wherein~$T_{ijk}$ is given
        by~\eqref{eq:overall_approx}, and each of the components has a single
        latent factor~$R_{(\cdot)}=1$.
\end{enumerate}

Models were trained using the Stochastic Gradient Descent
method~(SGD) of the momentum variant, with decreasing time-steps.
In all the approximations 1-5, the bias terms were taken to be identical.
Specifically, they were not trained by SGD but rather chosen, before the SGD
simulations, by~\eqref{eq:bias_terms}.
The parallel factors were regularized using the~$L^2$ norm, using nine-fold
cross-validation to pick the regularization parameter, and $25$-fold
cross-validation to measure performance of the best configuration.

\subsection{\label{ssec:datasets}The Datasets}

The MovieLens Dataset~\cite{MovieLens-Dataset} contains a million user-ratings
of movies on a scale of one to five.
Ratings of~$4$ and~$5$ were considered to be positive events, and lower ratings
as negative events.
Overall,~$424928$ negative and~$575281$ positive rating events were considered.
The three factors we consider are those of item, user and hour of week
(totaling~$168$ bins).

The Fannie Mae Single-Family Loan Performance
dataset~\cite{FannieMae-DataSet} is a publicly available dataset which, at the
time of submission, holds fixed rate prime mortgage acquisition and performance
data, at monthly resolution, for the period from January 1999 till June 2013,
including.
Only first-time home buyers whose loan purchase was buying or undefined were
considered.
The three factors chosen where credit-score, property location denoted by
property state and 3-digit zip code, and origination month.
We chose not to group or smooth different values of credit scores or time
periods longer than a month, so as not to make the prediction problem easier.
A mortgage was considered to have defaulted if delinquent more than 150 days over
the full period.
Non-default events were uniformly downsampled.
Overall,~$1197549$ non-default and~$876707$ default acquisition events were
considered.

\subsection{\label{ssec:num_results}Results}

\begin{table}
\begin{centering}
\begin{tabular}{|c|c||c|c|c|c|c|c|}
\hline 
\multicolumn{8}{|c}{MovieLens}\tabularnewline
\hline 
\multicolumn{2}{|c|}{Method} & AUC & $\Delta$AUC & L1 & $\Delta$L1 & L2 & $\Delta$L2\tabularnewline
\hline 
\hline 
1 & Bias only & $0.6494$ & $21$ & $0.4542$ & $6$ & $0.4712$ & $5$\tabularnewline
\hline 
2 & CP, $R=13$ & $0.7625$ & $36$ & $0.3387$ & $21$ & $0.4456$ & $22$\tabularnewline
\hline 
3 & best CP, $R=5$ & $0.7783$ & $56$ & $0.3470$ & $31$ & $0.4318$ & $26$\tabularnewline
\hline 
4 & primitive NCLF & $0.7817$ & $48$ & $0.3536$ & $39$ & $0.4283$ & $25$\tabularnewline
\hline 
5 & NCLF & $0.7920$ & $31$ & $0.3365$ & $23$ & $0.4256$ & $18$\tabularnewline
\hline 
\hline 
 & NCLF-best CP & $0.0137$ & $45$ & $0.0105$ & $27$ & $0.0062$ & $22$\tabularnewline
\hline 
\end{tabular}
\par\end{centering}

\caption{\label{tab:ML_num_results}Performance of the five approximations
as given in Section~\ref{ssec:approximations}, for the MovieLens
1M ratings dataset, obtained by 25-fold cross-validation. Columns
denoted by~$\Delta(\cdot)$ give sample standard errors, multiplied
by~$10^{4}$.
The last row gives the absolute difference of the CP with the best
rank~$R=5$ to the NCLF.}
\end{table}
\begin{table}
\begin{centering}
\begin{tabular}{|c|c||c|c|c|c|c|c|}
\hline 
\multicolumn{8}{|c}{Fannie Mae}\tabularnewline
\hline 
 & \multicolumn{1}{c|}{Method} & AUC & $\Delta$AUC & L1 & $\Delta$L1 & L2 & $\Delta$L2\tabularnewline
\hline 
1 & Bias only & $0.7689$ & $58$ & $0.3819$ & $7$ & $0.4329$ & $6$\tabularnewline
\hline 
2 & CP, $R=13$ & $0.8242$ & $38$ & $0.3028$ & $14$ & $0.4105$ & $25$\tabularnewline
\hline 
3 & best CP, $R=5$ & $0.8326$ & $55$ & $0.3062$ & $41$ & $0.4029$ & $33$\tabularnewline
\hline 
4 & primitive NCLF & $0.8447$ & $19$ & $0.3040$ & $12$ & $0.3954$ & $12$\tabularnewline
\hline 
5 & NCLF & $0.8462$ & $16$ & $0.3029$ & $9$ & $0.3942$ & $10$\tabularnewline
\hline 
\hline 
 & NCLF-best CP & $0.0136$ & $41$ & $0.0033$ & $30$ & $0.0087$ & $24$\tabularnewline
\hline 
\end{tabular}
\par\end{centering}

\caption{\label{tab:FM_num_results}Same as Table~\ref{tab:ML_num_results},
for the Fannie Mae dataset.}
\end{table}

Cross-validation performance of the five approximations of~\ref{ssec:approximations}
applied to the MovieLens dataset is given in Table~\ref{tab:ML_num_results}, and
their performance over the Fannie Mae dataset is given in
Table~\ref{tab:FM_num_results}.
In both cases, we see that the NCLF models considerably outperforms the standard
CP model of the same latent dimension~$13$, and significantly outperforms CP
models of lower dimensions, as measured by all metrics: AUC,~$L_1$ error
and~$L_2$ error.

The numerical experiments therefore strongly corroborate
Conjecture~\ref{conj:noncom}, at least for these datasets and with the SGD
numerical method - under these assumptions, non-commutative latent factors
outperform the standard CP.

Additionally, there is weak evidence that the proposed NCLF mildly outperforms
the ``primitive NCLF'' model~\eqref{eq:mu_only_parafac}, meaning that applying
the symmetrizing operators of
Section~\ref{ssec:approximating_S_and_J} (thereby approximating the two 
components~$\jacobi_{23-},\jacobi_{31-}$)
provides an improved approximation.

\section{\label{sec:discussion}Discussion and Future Directions}

In this study, we develop a novel tensor-completion method for three-way arrays,
which is both diagonal and built upon non-commutative latent factors.
In order to do this, we apply symmetrizing operations on the simplest
non-commutative purely trilinear operation we could find - that of three-matrix
product on a two-dimensional space.
We test our model and numerical method on a binary-response 
supervised-learning problem from two publicly-available datasets, finding that
it outperforms the CP model.

The specific application we are interested in is modeling sparse, large-scale
three-way relations in the supervised-learning setting, i.e., in three-way
CF problems.
However, we find no apriori reason that this model may not be extended to a
broader setting.
Some future avenues for research include:
\begin{enumerate}
    \item {\bf Unsupervised learning:} An interesting question is if and how
        much a non-commutative model may be used to discover non-commutative
        patterns in three-way-relation data.
        The intuitions leading to its development in Section~\ref{sec:intuitive}
        should still apply.
    \item {\bf (Dense) Tensor Factorization:}
        A possible future direction may be the analysis of
        this model in the context of tensor-factorization - i.e., of
        approximation a full tensor with no missing values.
        We note that in this setting there are Fourier-based generalizations of
        the SVD~\cite{kilmer-fft3rd-2011} in addition to the HOSVD of
        Delathauwer et al., and a comparison of the three options may be
        interesting.
    \item {\bf Extension to Quaternions:} The space~$(\CPerp,\mu)$ is in fact
        a two-dimensional subspace of the ring of quaternions.
        One may consider applying the symmetrizing
        operators~\eqref{eq:cuboid_symmetrizers} on three-quaternion products
        instead of on~$\mu$ - in fact, this was the original direction
        of this work.
        The resulting approximation might be more expressive than NCLF, but
        have a double latent dimension, and so be more likely to overfit.
        Nevertheless, in a world where the volume of data keeps increasing, such
        an extension might some day prove superior.
\end{enumerate}

In summary, Non Commuting Latent Factors present a simple, scalable extension
of the CP model which outperforms it on the two datasets tried.


\end{document}